\documentclass{article}

\usepackage{arxiv}

\usepackage[utf8]{inputenc} % allow utf-8 input
\usepackage[T1]{fontenc}    % use 8-bit T1 fonts
\usepackage{hyperref}       % hyperlinks
\usepackage{orcidlink}      % orcidlinks
\usepackage{url}            % simple URL typesetting
\usepackage{booktabs}       % professional-quality tables
\usepackage{amsfonts}       % blackboard math symbols
\usepackage{nicefrac}       % compact symbols for 1/2, etc.
\usepackage{microtype}      % microtypography
\usepackage{lipsum}		% Can be removed after putting your text content
\usepackage{graphicx}
\usepackage{cite} 
\usepackage{doi}
\usepackage{float}
\usepackage{amsmath}

\title{Stability of Ranking-dependent Pair-wise Comparison Patterns in the Analytic Hierarchy Process}

\author{{ \orcidlink{0000-0002-0821-4877} Vitaliy Tsyganok} \\
	Institute for Information Recording\\
	National Academy of Sciences of Ukraine\\
	Kyiv, Ukraine 03113 \\
	\texttt{tsyganok@ipri.kiev.ua} \\
	\And
	{ \orcidlink{0000-0001-7191-5636} Sergii Kadenko} \thanks{Corresponding author}\\
	Institute for Information Recording\\
	National Academy of Sciences of Ukraine\\
	Kyiv, Ukraine 03113 \\
	\texttt{seriga2009@gmail.com} \\
    \And
	{ \orcidlink{0000-0003-2569-2026} Oleh Andriichuk} \\
	Institute for Information Recording\\
	National Academy of Sciences of Ukraine\\
	Kyiv, Ukraine 03113 \\
	\texttt{andreychuck@ukr.net} \\
}

\renewcommand{\shorttitle}{\textit{Tsyganok, Kadenko, Andriichuk} Stability of Ranking-dependent PC Patterns in AHP}

\hypersetup{
pdftitle={Stability of Ranking-dependent PC Patterns in AHP},
pdfsubject={SC.AI},
pdfauthor={Vitaliy V.~Tsyganok, Sergii V.~Kadenko, Oleh V.~Andriichuk},
pdfkeywords={analytic hierarchy process, best-worst method, expert estimate, incomplete pair-wise comparison matrix, ranking, simulation, genetic algorithm},
}

\begin{document}
\maketitle

\begin{abstract}
	The paper addresses several ranking-dependent decision support methods. Ordinal information on compared objects can be used to improve the quality of expert data during estimation and help reduce the number of comparisons that the experts need to perform. In the paper we compare three incomplete ranking-dependent pair-wise comparison patterns which can be used in the Analytic Hierarchy Process – Best-worst method, Best-Second Best (Top 2) method, and the original maximum difference method. The first two comparison patterns (and respective methods) are incomplete, while the third can be a complete one. We determine conditions under which these three methods can be compared in terms of stability to expert errors. We also present the results of a simulation-type experiment, in which the three methods are compared. The research allows us to define the most stable incomplete rankingdependent pair-wise comparison pattern and reduce the number of comparisons without loss of credibility of expert session results. The research contributes to algorithmic, cognitive, and applied aspects of decision support in uncertain environments.
\end{abstract}

% keywords can be removed

\section{Introduction}
The present research addresses cognitive and algorithmic aspects of knowledge transfer and decision support. Particularly, our research of specific pair-wise comparison (PC) patterns in the Analytic Hierarchy Process (AHP)~\cite{Saaty1980} and related methods is intended to improve the quality of expert data during estimation and, potentially, reduce the number of comparisons that the experts need to perform.
In order to obtain relative weights of n objects in AHP, an expert or respondent needs to perform from $n$-1 to $n$($n$-1)/2 PCs. As the number of compared objects grows, the task becomes more labor-intensive. In order to simplify the process and reduce the computational complexity of AHP, while preserving sufficient levels of credibility and redundancy of expert data, different incomplete PC methods have been developed. Incomplete PC methods have been covered in multiple publications (such as~\cite{Wedley2009, Kulakowski2020, Szadoczki2022}). PC methods taking rough ranking of compared alternatives into consideration represent a separate group among them. A series of recent publications (including~\cite{AndriichukKadenko2022} ) addresses these particular methods.
While comparisons of incomplete PC methods in terms of stability to expert errors have been outlined in several studies (such as~\cite{Szadoczki2022}), a separate comparative analysis of ranking-dependent PC methods has not been conducted yet.
Algorithms for such comparative studies of PC methods as to their stability have been developed quite a long time ago~\cite{Tsyganok2010}. AHP and related methods are mostly applied in weakly structured subject domains also called uncertain environments. These are influenced by multiple intangible factors, for which there are no benchmark values or measurement units. Engaging actual experts in such experiments aiming to compare different methods requires considerable resources. Therefore, it makes sense to focus on simulation-based modeling of the expert estimation process.
We should note that in~\cite{Szadoczki2023} PC methods, using different amounts of ordinal information (Best-worst, Best-Random, 3(-quasi)-regular-graphs, others), have been com-pared in terms of both ordinal and cardinal stability. The issues of legitimacy of such “mixed” comparisons became relevant before the publication of article~\cite{Szadoczki2023} and have remained the subject of discussion ever since.
Our main objective is to compare the three chosen ranking-dependent PC methods: Best-worst~\cite{Rezaei2015, Rezaei2016}, Top 2 (Best-Second Best)~\cite{Szadoczki2023}, and the original max difference method~\cite{Andriichuk2024}. Additionally, we would like to validate the max difference method by comparing it with other ranking-dependent ones. However, before conducting a simulation-type comparison of the listed methods, we need to define the conditions, under which the three listed methods can be compared among themselves (i.e., conditions allowing us to consider the results of the methods’ experimental comparison according to stability criterion credible enough). So, in further sections we will describe comparison methodology and then move on to the simulation experiment and its result.

\section{Materials and Methods}
\subsection{Methods overview}
Let us start with a quick overview of the methods under consideration. If the rough ranking of compared objects is available, then the Best-worst method suggests comparing all the objects only to the best and the worst one in the set (according to the ranking). Respectively, the Top 2 method suggests comparing all the objects only with the best and the second-best ones in the ranking. PC structures of Best-worst and Top 2 methods can be easily illustrated by graphs. The looks of such graphs for the case when 7 objects are compared is shown on Figure 1. Sizes of nodes indicate the ranks of respective objects.
\begin{figure} [H]
    \includegraphics[width=0.5\linewidth]{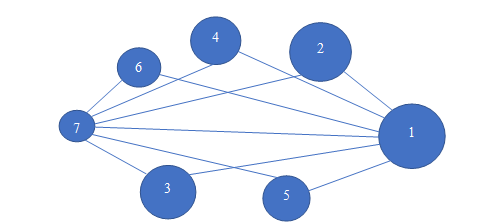} 
    \includegraphics[width=0.5\linewidth]{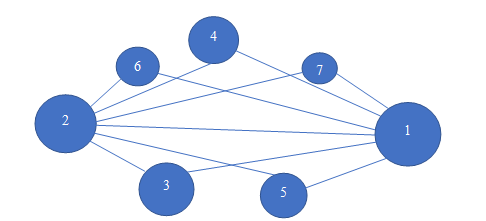}
    \label{fig:placeholder}
    \caption{Graphs of PC structures for 7 objects: Best-worst method (left) and Top 2 method (right)}
\end{figure}  

In both Best-worst and Top 2 methods each object is compared to only two objects in the ranking (either the best and the worst or the best and the second best). Therefore, both methods are incomplete ones (not all PCM cells contain independent PC values), and the number of comparisons amounts to ($n$-1)+(($n$-1)-1)=2$n$-3 PCs. If the objects in the ranking are numbered from best to worst, then the respective in-complete PCM for Best-worst (left) and Top 2 (right) methods look as follows:
\begin{figure}[H]
    \centering
    \includegraphics[width=1\linewidth]{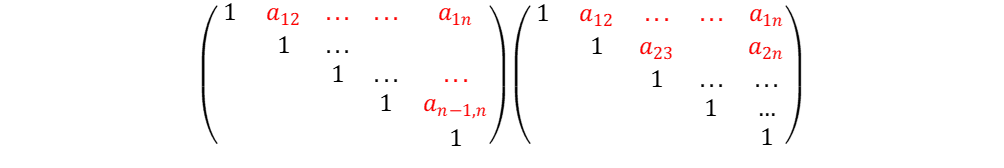}
\end{figure}

In Best-worst method only the first row and the last column of a PCM are filled, while in Top 2 method we need to fill only the first two PCM rows.
Max difference method envisions a certain sequence of PC, defined by the rough ranking of the objects. Let’s assume, the objects are numbered according to their rank order: $a_1$>$a_2$>...>$a_n$, where $a_i$ is the object with rank and number $i$, $i$=(1,$n$) and n is the total number of objects. In this case the sequence of comparisons (i.e., of object pairs presented to the expert), ensuring the highest relevance and consistency of the results~\cite{Andriichuk2024} is as follows:

\begin{itemize}
  \item batch 1: ($a_1$,$a_n$) (ranks differ by ($n$-1));
  \item batch 2: ($a_1$,$a_{(n-1)}$) or ($a_2$,$a_n$)  (ranks differ by ($n$-2));
  \item batch 3: ($a_1$,$a_{(n-2)}$) or ($a_2$,$a_{(n-1)}$) or ($a_3$,$a_n$) (ranks differ by ($n$-3));
  \item … ;
  \item batch ($n$-1): ($a_1$,$a_2$) or ($a_2$,$a_3$) or … or ($a_{(n-1)}$,$a_n$) (ranks differ by 1). 
\end{itemize}

The respective PCM is filled as shown below:

\begin{figure}[H]
    \centering
    \includegraphics[width=1\linewidth]{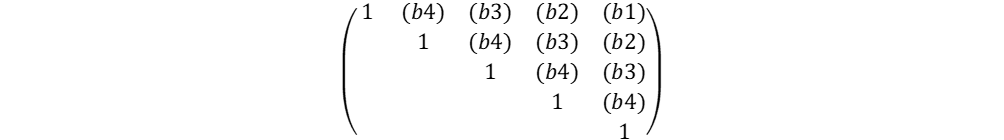}
    \caption{The pattern of PCM filling in “batches” (b1-b4) in the max difference method}
    \label{fig:placeholder}
\end{figure}

\subsection{Ensuring equal numbers of comparisons in incomplete PC patterns}

In~\cite{Andriichuk2024} it is shown that it is not necessary to perform all PCs in max difference method. All PCs within the same batch are “equally important”, so we can confine ourselves to PCs which are required for obtaining a connected PC structure (graph). 
If n is an odd number, then in order for the PC graph to become connected it is enough to perform $[n/2]$ or ($n$-1)/2 PC batches + 1 PC from the next batch, i.e. batch number (($n$-1)/2)+1. This might be the PC of objects ($a_1$,$a_{((n-1)/2+1)}$), i.e. the 1st and the “central” element in the ranking. Likewise, it can be the PC of the “central” element and the last one ($a_{((n-1)/2+1}$),$a_n$), as all PCs within a batch are equivalent, and objects can be numbered from best to worst or vice versa. The point is to include the “central” ranking element into the PC structure, so that it becomes connected. 
If $n$ is an even number, then there will be two “central” elements in the ranking. Therefore, in order to obtain a connected comparison structure, we need to perform $(n/2)-1$ PC batches + 2 PC from batch number $(n/2)$: ($a_1$,$a_{(n/2)}$), ($a_{(n/2+1)}$,$a_n$). An example of obtaining a connected comparison graph for 7 objects is shown on Figure 3.

\begin{figure}[H]
    \centering
    \includegraphics[width=0.5\linewidth]{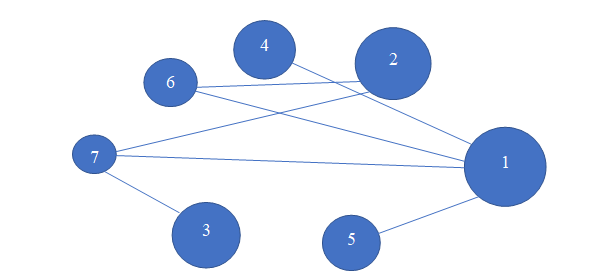}
    \caption{A connected set of PC of 7 objects in max difference method}
    \label{fig:placeholder}
\end{figure}

In this example batch 1 includes a single PC ($a_1$,$a_7$), batch 2 – PCs {($a_1$,$a_6$), ($a_2$,$a_7)$}, batch 3 – PCs {($a_1$,$a_5$), ($a_2$,$a_6$), ($a_3$,$a_7)$}, batch 4 – PCs {($a_1$,$a_4$), ($a_2$,$a_5$), ($a_3$,$a_6$), ($a_4$,$a_7$)} and so on. Connectivity is achieved after PC ($a_1$,$a_4$) or, alternatively, ($a_4$,$a_7$).
By definition, the number of the batch $i$ equals the total number of PCs in it. Therefore, the total number of PCs in batches from 1 to $i$ equals the sum of the respective arithmetic progression: $N_i=i(i+1)/2$. That is, in order to obtain a connected set of PCs of n objects, compared using the max difference method, we need to perform the following number of comparisons (1):

\begin{equation}
	N_{\mathrm{conn}}=
    \begin{cases}
    \left(\dfrac{n-1}{4}\left(\dfrac{n-1}{2}+1\right)\right)+1, & n=2k-1,\quad k\in\mathbb{N}, \\[0.8em]
    \left(\dfrac{n}{4}\left(\dfrac{n}{2}-1\right)\right)+2, & n=2k,\quad k\in\mathbb{N}.
    \end{cases}
\end{equation}

In order to compare $n$ objects among themselves using either Best-worst or Top 2 method, we need to perform (2$n$-3) PC (as explained earlier). The max difference method, in its turn, is a complete one. So, to place the listed 3 methods (Best-worst, Top 2, max difference) in equal conditions and be able to compare their stability to expert errors, we have to ensure that the number of PCs in all 3 methods is equal. Therefore, the following condition (2) should be fulfilled:

\begin{equation}
	N_{\mathrm{conn}} \le 2n - 3
\end{equation}

Having solved the respective square inequalities, we can deduce that condition (2) is fulfilled when $3\le n\le 14$. 

Minimum numbers of PCs, required for calculation of priorities using Best-worst, Top 2, and max difference methods, as functions of n are shown on Figure 4. For Best-worst and Top 2 methods this dependence function is a linear one ($N$=$O$($n$)), i.e. its graph is a straight line. For max difference method, this dependence function is a square one ($N_{conn}$=$O$($n^2$)). Its graphs for odd and even $n$ are parabolic. Therefore, when the number of objects lies within the limits $3\le n\le 14$, in order to compare the 3 methods, we should generate (2$n$-3) PCs in max difference method, according to the sequence suggested by the batches. That is, after the “connectivity number” of comparisons $N_{conn}$ is achieved, we should keep generating PCs until their number reaches (2$n$-3).

\begin{figure}[H]
    \centering
    \includegraphics[width=0.75\linewidth]{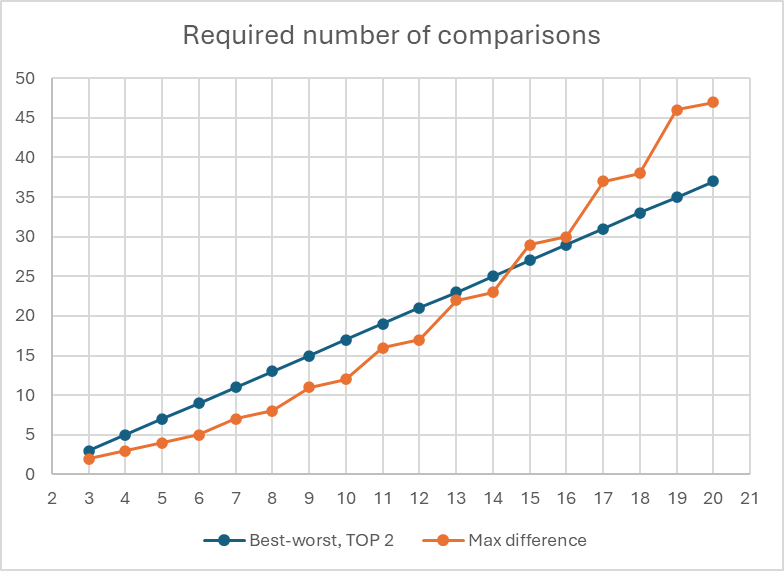}
    \caption{The number of PCs, required for priority vector calculation in Best-worst, Top 2, and max difference methods}
    \label{fig:placeholder}
\end{figure}

If $n$>14, then condition (2) in max difference method cannot be fulfilled. In this case we propose to start generating PC in max difference method with a basic PC set, suggested in~\cite{Andriichuk2024}, that is a bi-partite spanning tree graph, in which $a_1$ (the best object) is compared to all objects from the 2nd half of the ranking ($a_{([n/2]+1)}$,…,$a_n$), while $a_n$  (the worst object) is compared to all objects from the 1st half of the ranking ($a_1$,…,$a_{[n/2]}$). Diameter of this graph equals 3 and does not depend on $n$, while its edges represent comparisons from first batches, that is comparisons of objects, whose ranks are most distant. These properties of the graph ensure stability of this initial PC structure to expert errors~\cite{Kulakowski2020}. An example of such a spanning tree graph for 7 objects is shown on Figure 5.

\begin{figure}[H]
    \centering
    \includegraphics[width=0.5\linewidth]{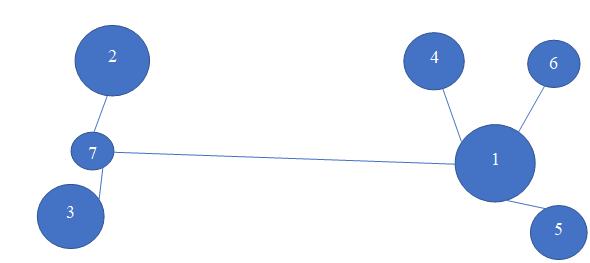}
    \caption{Spanning tree graph, representing the minimum connected PC set for 7 objects, built using max difference method}
    \label{fig:placeholder}
\end{figure}

In fact, the only difference between graphs shown on Figure 5 and Figure 3 is the edge corresponding to PC ($a_2$,$a_6$), which makes PC set on Figure 3 redundant and is missing on Figure 5.
Once this initial connected set of ($n$-1) PC is built, we should add PCs to the set, following the order of batches, until the number of PCs reaches (2$n$-3). A basic flow chart of PC generation algorithm for max difference method can be found in the Appendix.
Once the necessary number of PCs is generated, we can compare the stability of max difference, Best-worst, and Top 2 methods, following the procedure outlined in the next section.

\section{Results}
\subsection{Experiment outline}
Experimental research of a PC method’s stability might include the following conceptual phases:
\begin{enumerate}
  \item [1.] 	Generate a set of relative weights of compared objects ($w_1$,…,$w_n$), i.e., priority vector
  \item [2.]	Generate a PC matrix (PCM) of judgments based on these relative object weights {$a_{ij}$=$w_i$/$w_j$; $i$,$j$=1..$n$}
  \item [3.] Generate an incomplete PCM based on the complete one, according to a given method (max difference, Best-worst, Top 2)
  \item [4.] Fluctuate the generated PCM: $a_{ij}' = a_{ij}(1+\varepsilon)^{\pm 1}, 
  \qquad 0<\varepsilon<1,\quad i,j=1,\dots,n$.
  \item [5.] Calculate priorities ($w_1'$,…,$w_n'$) using some given method suitable for incomplete PCM~\cite{BozokiFulopRonyai2010} (least squares~\cite{BozokiTsyganok2019}, spanning tree enumeration~\cite{Tsyganok2010,Kadenko2021}, other) based on the fluctuated PCM. In our particular experiment we decided to use the combinatorial spanning tree enumeration method~\cite{Kadenko2021,Tsyganok2010} to calculate priorities. This choice is dictated by the method’s efficiency for incomplete PCM and by sophisticated incomplete PC patterns (not just rows or columns of PCM). Each of the three ranking-dependent PC methods we analyze involves such patterns.
  \item [6.] Calculate the difference between initial and calculated priority vectors (i.e. priority deviation) $\Delta$. This difference can be represented by any of the indicators (Euclidean distance, Kendall’s Tau~\cite{Szadoczki2023} or similar ones).
\end{enumerate}

Maximum deviation of the weight vector, obtained based on the fluctuated PCM, from the initial vector $\Delta$, may be used as the indicator of a method’s stability to expert errors. According to this indicator, different PC methods are compared with each other. In the current experimental research, we suggest using maximum relative deviation of priority vector coordinates from initial values (3) as a stability indicator:

\begin{equation}
\Delta = \left(\max_i \frac{\max\!\left(w_i, w_i'\right)}{\min\!\left(w_i, w_i'\right)} - 1\right) \cdot 100\%
\end{equation}

In order to search for maximum deviations of priority vector coordinates, we implement the genetic algorithm (GA)~\cite{Holland1975}, allowing us to conduct a directed search among fluctuated PCMs until the value of $\Delta$ stabilizes. In terms of the GA, various fluctuated PCMs represent a population of individuals, to which cross-breeding and mutations are applied (in order to produce new PCM “generations”), while $\Delta$ is the fitness function of the GA. With each new generation of PCMs $\Delta$ increases. If it stops increasing and does not change significantly during a predefined number of learning epochs, then the algorithm stops and the value of $\Delta$ is considered maximum.

Section 2 shows, that in our current research we have found a kind of a threshold dimensionality value $n$=14. In a previous simulation-type study similar to the current one~\cite{Kulakowski2020} dimensionality of analyzed incomplete PCM ranged from $n$=4 to $n$=24. At the same time, according to psychophysiological constraints of human mind, outlined in~\cite{Miller1956} and elaborated in~\cite{SaatyOzdemir2003}, an average expert is capable of ranking and rating only $n$=(7$\pm$ 2) objects (with some degree of credibility). Therefore, we should note that actual estimation problems are, usually, limited to these dimensionalities. So, at present, our simulation-type experiment is focused on modeling of individual estimation problems where the number of compared objects is $n$=(7$\pm$ 2).

The software tool used to conduct the simulation-type experiment described in the previous subsection is similar to the one outlined in~\cite{Kadenko2021} (Figure 6). It allows the user to select a particular PC method, set deviation level, fluctuation step value, number of individuals in a population, mutation probability, and the number of generations after which the algorithm stops if fitness function value does not improve.

On Figure 6 we show an example if an experiment where the initial PCM fluctuation (deviation) level $\varepsilon$ is set at 0.01\%. After that it is gradually increased, starting with a “step” of 0.01\%. The number of fluctuated PCMs in the population is 150. After fitness function is calculated a new generation of the “fittest” PCMs is created using cross-breeding and mutation~\cite{Holland1975}. If fitness function $\Delta$ remains stable for 500 generations in a row, then the algorithm moves on to larger fluctuation, gradually increasing the increment value. Within the context of the present study, we’ve decided to set the maximum fluctuation (hypothetical “expert error”) value $\varepsilon$=30\% (as shown on Figures 7-12 below).

\begin{figure}[H]
    \centering
    \includegraphics[width=0.75\linewidth]{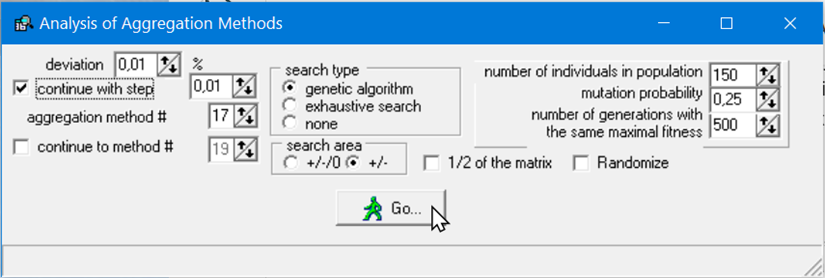}
    \caption{Screenshot of the software tool used for simulations}
    \label{fig:placeholder}
\end{figure}

Figure 7 shows a typical dependence of priority deviation $\Delta$ (vertical axis) on judgment perturbation $\varepsilon$ (horizontal axis) in the case when $n$=7. The exact priority vector used in this example is (9; 5; 2.1; 1.5; 1.3; 1.1; 1).

\begin{figure}[H]
    \centering
    \includegraphics[width=0.75\linewidth]{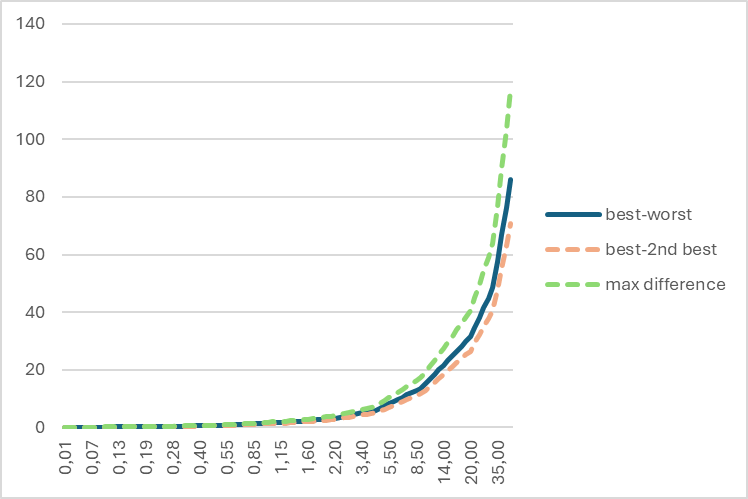}
    \caption{Example of dependance of $\Delta$ on $\varepsilon$ for Best-worst, Top 2, and max difference methods when $n$=7; $W$=(9; 5; 2.1; 1.5; 1.3; 1.1; 1) (PCM fluctuation does not take the ranks of compared objects into account)}
    \label{fig:placeholder}
\end{figure}

We should note that, in the example on Figure 7 dependence of expert error size on rank differences of compared objects is not taken into account. That is, fluctuation of PCM elements was performed irrespectively of rank differences ($a_{ij}'$=$a_{ij}$(1+$\varepsilon)^{(\pm {1})};0<\varepsilon$<1,$i$,$j$=1..$n$). As we can see, under this scenario, best-second best (TOP 2) method displays a certain advantage over Best-worst and max difference methods in terms of stability to expert errors. However, this experiment does not allow us to consider all the necessary information during the simulation process.

\subsection{Modifying the PCM fluctuation process}
Before moving on to further numeric results, we would like to dwell on PCM fluctuation process, intended to simulate expert errors (step 4 of the experiment algorithm from subsection 3.1). Analysis of results obtained in~\cite{Andriichuk2024} indicates that adequacy of expert estimates is, indeed, significantly influenced by the order in which pairs of objects are presented to the expert for comparison. Particularly, PCs belonging to batches with smaller numbers (i.e. comparisons of more distant objects) are more credible than PCs from batches with larger numbers (i.e. comparisons of closer objects). Benefits of max difference method include higher adequacy of priorities and better consistency (Saaty’s consistency ratio (CR)~\cite{Saaty1980}) of judgments~\cite{Andriichuk2024}. Therefore, in our view, it is irrelevant and insufficient to apply simple multiplicative fluctuations (such as the ones described in a somewhat similar experiments in~\cite{Kadenko2021,Tsyganok2010} to PCM during comparison of the three methods), as they cannot convey all the information about estimation process. That is, while fluctuating PCM elements (step 4 of the simulation procedure set forth in subsection 3.1), we need to take the numbers of specific batches these elements belong to (i.e. the distance between the respective compared objects in the ranking) into account. Based on the outcomes of our previous research~\cite{Andriichuk2024}, we assume that comparisons from batches with smaller numbers are less prone to errors than comparisons from subsequent batches. During simulation process this can be achieved if we modify ordinary multiplicative fluctuation formulas in such a way that fluctuation of a particular PCM element depends on the number of the batch the respective comparison belongs to. By definition of the batches, the number of a batch k is determined by the distance between the respective compared objects in the ranking:  $k=n-|i-j|$. Thus, the fluctuation of a specific judgment (PCM element) should be inversely proportionate to the distance between the compared objects in the ranking, i.e. $|i-j|$.
At the same time, comparing more distant objects helps overcome the famous anchoring bias, described in~\cite{TverskyKahneman1974,Ni2019}. The Best-worst method~\cite{Rezaei2015,Rezaei2016}, in a way, mitigates the bias. Furthermore, TOPSIS and the related family of multi-criteria decision support methods~\cite{HwangYoon1981} also rely on “calibration” of decision alternatives based on the best and the worst ideal options.
So, within the context of this paper we need to reiterate that both our previous research~\cite{Andriichuk2024} and earlier studies (such as~\cite{Rezaei2015,Rezaei2016,TverskyKahneman1974}) imply that expert errors depend on the ordinal distance between compared objects (i.e. the difference between their ranks). Finding specific functions most accurately describing the exact character of this dependence can be a subject of a separate further research, while in the current study we have decided to use the following generalized fluctuation formula (4) for our simulation experiment:

\begin{equation}
	a_{ij}' = a_{ij}
    \left(1 + \alpha\varepsilon + \frac{(1-\alpha)\varepsilon}{|i-j|}\right)^{\pm 1},
    \qquad i,j=1,\dots,n,\quad 0<\varepsilon<1,\quad 0\le \alpha\le 1.
\end{equation}

In (4) $\alpha$ defines the degree of dependence of expert error upon distance between compared objects in the ranking. If $\alpha=0$ then fluctuation heavily depends on it, while if $\alpha$=1 then fluctuation does not depend on rank difference at all. In terms of these considerations, Figure 7 illustrates the case when $\alpha$=1. During the simulation, the power indicator in formula (4) (1 or (-1)) is chosen at random. That is, if an expert’s PC of objects with numbers $i$ and $j$ is smaller than the actual ratio of priorities $a_{ij}$=($w_i$/$w_j$), then the power indicator in formula (4) equals (-1), that is to simulate the expert error, the respective element of the initial PCM $a_{ij}$ should be divided by expression ($1+\alpha\varepsilon+(1-\alpha)\varepsilon/|i-j|$). If the estimate is larger than the actual value, then $a_{ij}$ should be multiplied by this expression.

\subsection{Further Analysis and Interpretation of Results}
Experiments show that if $\alpha$=1 (i.e. expert errors do not depend on ordinal distance between compared objects) then max difference method is less stable than Top 2 and Best-worst methods. An example is shown on Figure 7.
If we assume that in (4) $\alpha$=0.5 (i.e. expert errors moderately depend on ordinal difference between compared objects) and use the same exact priority vector (9; 5; 2.1; 1.5; 1.3; 1.1; 1) in our experiment, then dependence of maximum deviation $\Delta$ (vertical axis) on fluctuation level $\varepsilon$ looks as shown on Figure 8. As we can see, Top 2 (Best – 2nd best) method remains the most stable one, but the difference in stability of Best-worst and max difference methods becomes smaller.

\begin{figure}[H]
    \centering
    \includegraphics[width=0.75\linewidth]{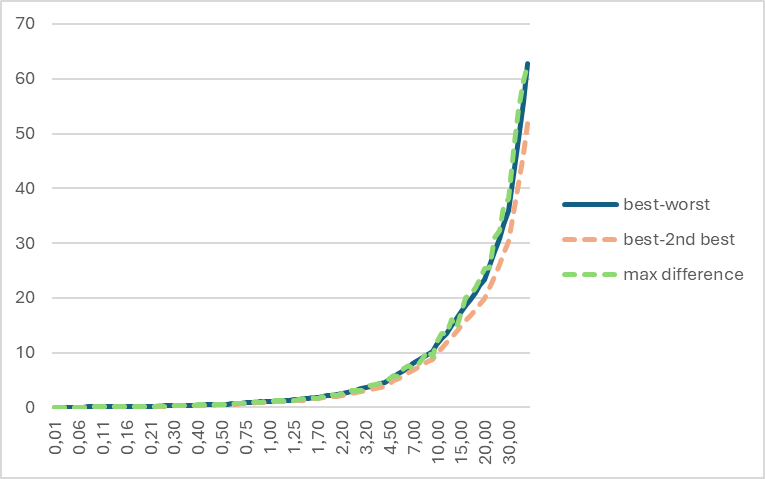}
    \caption{Example of dependance of $\Delta$ on $\varepsilon$ for Best-worst, Top 2, and max difference methods when $n$=7; $W$=(9; 5; 2.1; 1.5; 1.3; 1.1; 1); $\alpha$=0.5 }
    \label{fig:placeholder}
\end{figure}

Finally, if we assume that in (4) $\alpha$=0 (i.e. expert errors strongly depend on ordinal difference between compared objects) and use the same exact priority vector (9; 5; 2.1; 1.5; 1.3; 1.1; 1) in the experiment, then dependence of maximum deviation $\Delta$ (vertical axis) on fluctuation level $\varepsilon$ looks as shown on Figure 9. As we can see, under this scenario max difference method becomes the most stable one.
The trend shown on Figures 7-9 does not look exactly the same for all priority vectors. That is, if we use another exact priority vector in the experiment, then max difference method displays better results in terms of stability under larger values of $\alpha$.

\begin{figure}[H]
    \centering
    \includegraphics[width=0.75\linewidth]{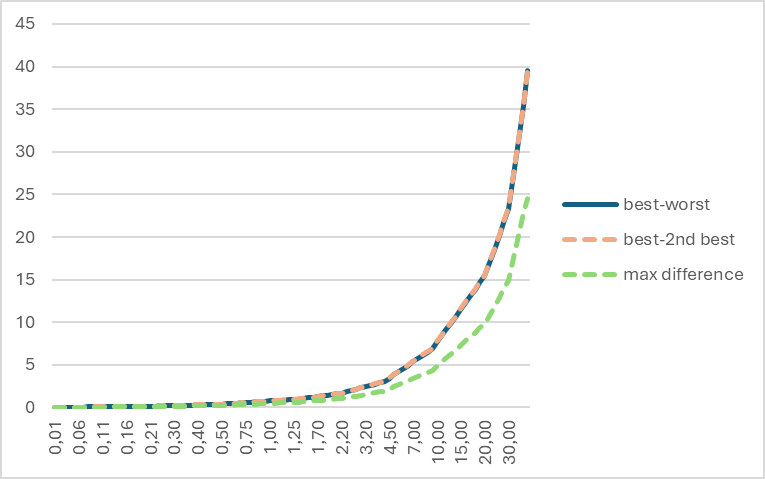}
    \caption{Example of dependance of $\Delta$ on $\varepsilon$ for Best-worst, Top 2, and max difference methods when $n$=7; $W$=(9; 5; 2.1; 1.5; 1.3; 1.1; 1); $\alpha$=0 }
    \label{fig:placeholder}
\end{figure}

Let us consider one more example, where exact priorities used in the simulation experiment, fall within a smaller range and include identical weights: $W$=(4; 3,4; 3; 3; 2,7; 2.5; 2.5). Figure 10 shows the behavior of $\Delta$ for the three methods when $\alpha$=1 (i.e. expert errors do not depend on ordinal difference between compared objects). As we can see, Top 2 method is the leader in terms of stability, followed by Best-worst and max different methods, just like in the case of the previous priority vector (Figure 7).

\begin{figure}[H]
    \centering
    \includegraphics[width=0.75\linewidth]{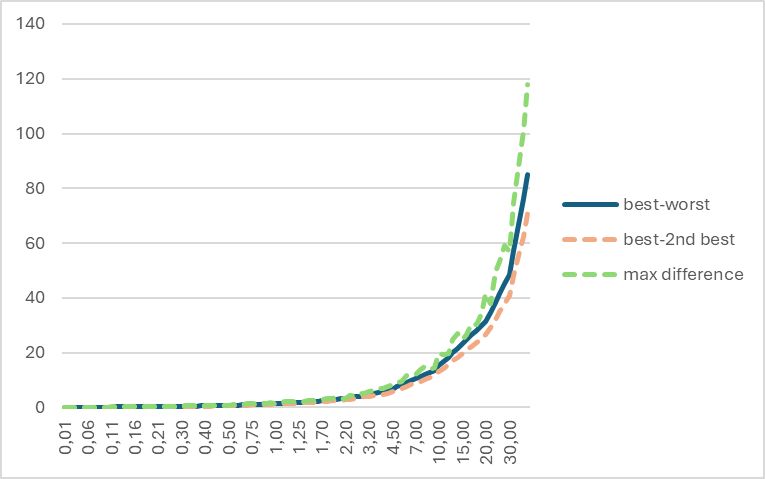}
    \caption{Example of dependence of $\Delta$ on $\varepsilon$ for Best-worst, Top 2, and max difference methods when $n$=7; $W$=(4; 3,4; 3; 3; 2,7; 2.5; 2,5); $\alpha$=1 }
    \label{fig:placeholder}
\end{figure}

\begin{figure}[H]
    \centering
    \includegraphics[width=0.75\linewidth]{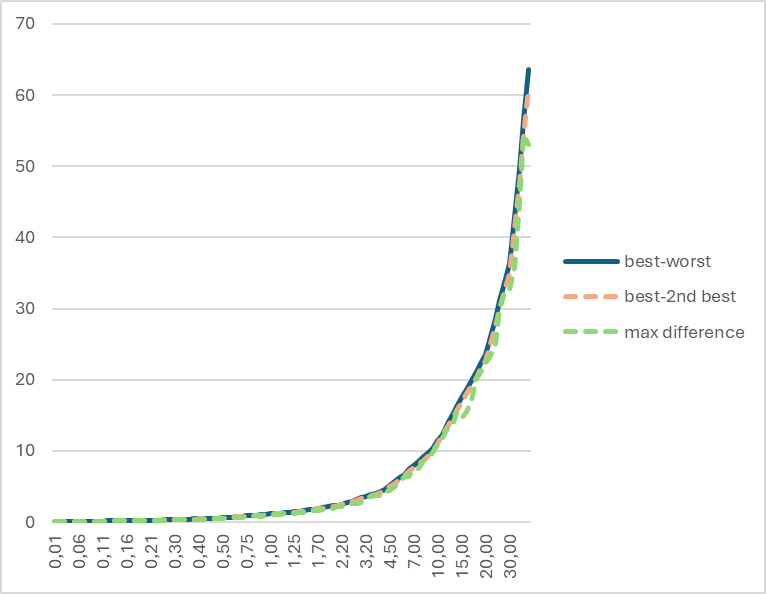}
    \caption{Example of dependence of $\Delta$ on $\varepsilon$  for Best-worst, Top 2, and max difference methods when $n$=7; $W$=(4; 3,4; 3; 3; 2,7; 2.5; 2,5); $\alpha$=0.5 }
    \label{fig:placeholder}
\end{figure}

\begin{figure}[H]
    \centering
    \includegraphics[width=0.75\linewidth]{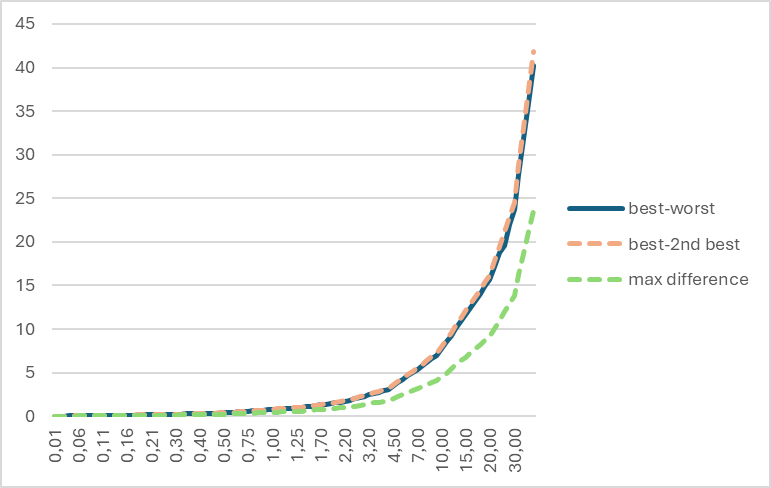}
    \caption{Example of dependence of $\Delta$ on $\varepsilon$  for Best-worst, Top 2, and max difference methods when $n$=7; $W$=(4; 3,4; 3; 3; 2,7; 2.5; 2,5); $\alpha$=0}
    \label{fig:placeholder}
\end{figure}

Figures 11 and 12, illustrate the cases when the exact priority vector $W$=(4; 3,4; 3; 3; 2,7; 2.5; 2,5) is used, while $\alpha$=0.5 and $\alpha$=0, respectively. It is worth noting that in case of this particular priority vector max difference method becomes the leader among the three in both cases.
Another trend worth noting is that with the decrease of $\alpha$, maximum deviation $\Delta$ also decreases in all three methods, irrespectively of specific priority values.

\section{Discussion and further research}
Results obtained thus far are based on the findings of our previous research~\cite{Andriichuk2024}, which lead us to assumption that experts tend to make larger errors when they compare objects that are closer to each other in the ranking. At the same time, related research in cognitive psychology by Stevens and Galanter~\cite{StevensGalanter1957} implies a slightly different trend. In their study, Stevens and Galanter analyzed errors made by respondents while evaluating objects according to 12 tangible criteria (such as loudness, weight, square, others).   
Research of Stevens and Galanter features estimates of evenly distributed values. In our examples (Figures 7-12) this even distribution requirement does not necessarily apply. However, if we accept the even distribution assumption (ranks of objects are evenly distributed anyway), then we might simulate the expert errors accordingly.

\begin{figure} [H]
    \centering
    \includegraphics[width=0.75\linewidth]{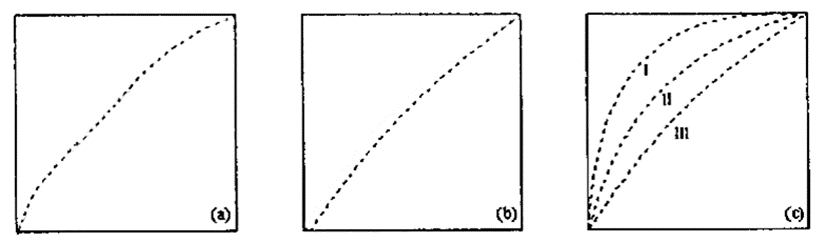}
    \caption{Respondents’ estimates of objects according to tangible criteria (steel core lengths (a), sound signal durations (b), and rectangle squares (c)), according to~\cite{ StevensGalanter1957}}
    \label{fig:placeholder}
\end{figure}

On Figure 13 charts (a) – (c) illustrate experiments of Stevens and Galanter, during which the respondents estimated objects according to respective tangible criteria, such as steel core length (a), sound signal duration (b), and rectangle square (c). True values of these characteristics of respective objects were evenly distributed along the diagonal of each square displayed on charts (a) – (c). 
As we can see from Figure 13, the closer to the middle of the ranking and/or range the estimated values lie, the larger the error is. Thus, we might want to simulate expert errors based on this “closeness to the middle” criterion:

\begin{equation}
a_{ij}'=
\begin{cases}
a_{ij}\left(1+\dfrac{\varepsilon}{
\max\!\left(\alpha,\min\!\left(\left|\left\lfloor \frac{n}{2}\right\rfloor+1-i\right|,\left|j-\left\lfloor \frac{n}{2}\right\rfloor-1\right|\right)\right)}\right)^{\pm 1},
& n=2k+1, \\[1.2em]
a_{ij}\left(1+\dfrac{\varepsilon}{
\max\!\left(\alpha,\min\!\left(\left|\frac{n}{2}-i\right|,\left|\frac{n}{2}+1-j\right|\right)\right)}\right)^{\pm 1},
& n=2k,
\end{cases}
\qquad k\in\mathbb{N},\quad 0<\alpha<1.
\end{equation}

In (5) $\alpha$ is an arbitrarily selected constant between 0 and 1, ensuring that the largest “noise” is added to the values, closest to the middle of the ranking. If $\alpha$=1 then a whole set of values from the middle of the ranking (for instance 3 out of 7, 4 out of 8) are perturbed with the same noise. If $\alpha$=0 then the noise is infinite. However, if 0<$\alpha$<1 then only the middle values (i.e. 1 value out of 2$k$+1 and 2 values out of 2$k$) are perturbed with maximum noise.
The problem is that in the general case actual estimated priority values are not evenly distributed (take, for instance, a priority vector like (9; 8.9; 8; 1.3; 1.2; 1.1; 1) or (9; 5; 3; 2; 1)). Therefore, in the context of our simulation experiment it might make sense to use the “middle of the range” value instead of the “middle of the ranking”. If priorities are ordered in the decreasing order, then:

\begin{equation}
    M=\frac{\lvert w_{\max}-w_{\min}\rvert}{2}
    =\frac{\lvert w_1-w_n\rvert}{2}
\end{equation}

If we use actual priority values instead of ranks in the perturbation formula above, then it might look as follows:

\begin{equation}
    a_{ij}' = a_{ij}
\left(
1 + \frac{\varepsilon}{
\max\!\left(\alpha, \min\!\left(\lvert w_i - M\rvert, \lvert w_j - M\rvert\right)\right)
}
\right)^{\pm 1},
\qquad n>2,\quad n\in\mathbb{N},\quad 0<\alpha<1.
\end{equation}

Introduction of $\alpha$ into this last modification of the perturbation formula allows us to reflect the assumption that the closer the objects’ weights are, the harder it is for the expert to estimate and compare them accurately.
All these considerations still do not provide us with a clear mechanism for simulation of specific pair-wise comparison sequences (such as “most distant object comparisons first”~\cite{Andriichuk2024} or other). 
Additionally, formulas for expert error simulation can reflect the size of expert errors from Figure 13 (c) based on the order in which the objects are presented to experts for estimation. For instance, based on rough estimate of expert error sizes (provided on  Figure 13 (c) and on Figure 13 (b) in Stevens and Galanter’s original article~\cite{StevensGalanter1957}), we can add the respective multiplier to the denominator of the fraction in formulas (5) and (7):

\begin{equation}
    a_{ij}'=
\begin{cases}
a_{ij}\left(1+\dfrac{\varepsilon}{
s\,\max\!\left(\alpha,\min\!\left(\left|\left\lfloor \frac{n}{2}\right\rfloor+1-i\right|,\left|j-\left\lfloor \frac{n}{2}\right\rfloor-1\right|\right)\right)
}\right)^{\pm 1},
& n=2k+1, \\[1.2em]
a_{ij}\left(1+\dfrac{\varepsilon}{
s\,\max\!\left(\alpha,\min\!\left(\left|\frac{n}{2}-i\right|,\left|\frac{n}{2}+1-j\right|\right)\right)
}\right)^{\pm 1},
& n=2k,
\end{cases}
\qquad k\in\mathbb{N},\quad 0<\alpha<1.
\end{equation}

\begin{equation}
    a_{ij}' = a_{ij}
\left(
1 + \frac{\varepsilon}{
s\,\max\!\left(\alpha,\min\!\left(\lvert w_i - M\rvert,\lvert w_j - M\rvert\right)\right)
}
\right)^{\pm 1},
\qquad n>2,\quad n\in\mathbb{N},\quad 0<\alpha<1.
\end{equation}

In (8) and (9) $s$ is the number of estimation sequence~\cite{Andriichuk2024,StevensGalanter1957}. In our simulation experiment, this would mean $s$=2 for best worst and Top 2 methods and $s$=3 for max difference method.
Thus, conducting of a simulation experiment with other priority fluctuation formulas (such as (8) and (9)) will be one of the key directions of our further studies. Other future research directions include: 

\begin{itemize}
    \item [1.] simulation-type comparative experiments involving larger numbers of compared objects ($n$>7);
    \item [2.] extrapolation of the approach to group decision-making cases;
    \item[3.] empirical definition of specific numbers of comparisons, allowing to achieve maximum stability of max diff method for different dimensionalities (similarly to~\cite{Wedley2009}).  
\end{itemize}

While in this paper we focused on algorithmic and cognitive aspects of decision support, omitting the applicational aspect, the approaches we set forth are intended for usage in specific weakly structured subject domains, such as defense and political environment modeling. It is possible to use them in combination with AI-powered technologies, for instance, as described in~\cite{Poudel2024}. We should note that when it comes to ad hoc decisions and choices in weakly-structured domains and changing environments, AI tools sometimes tend to fail decision-makers and/or provide misleading information as there might be no credible and relevant text corpora from which these tools could learn, not to mention ethical and legal issues~\cite{Cath2018} of AI-assisted decision-making. That is why expert data-based decision support methods, including those involving PCs, remain relevant.

\section{Conclusions}
In the present paper we have conducted an experimental comparative study of stability of several ranking-dependent PC methods (Best-worst, Top 2, max difference method) to expert errors. We have defined the conditions under which such an experiment can be performed and its results can be deemed valid. We have also suggested modifications of a typical simulation-type experiment required to compare the three listed methods in terms of stability. Particularly, based on experience of previous studies in the area, we have shown that accuracy of PCs is influenced by positions of compared objects in their rough ranking.
The experiment outlined in the current study is the first attempt to compare ranking-dependent PC methods without “mixing” them with complete and incomplete methods, which do not use ordinal preference information. In contrast to other comparative studies, where statistical approaches are used, we use GA for targeted search of maximum deviation $\Delta$ among all possible priority vectors for a given PCM fluctuation level $\varepsilon$.

Numeric results of the simulations, performed so far, speak in favor of the max difference method. That is, in comparison to Best-worst and Top 2 methods, the max difference method turns out to be more stable to expert errors. Another advantage of max difference method is that, in contrast to Top 2 and best-worst methods, it does not impose any specific configuration of incomplete preference structure (i.e. the number of comparisons can range from ($n$-1) to $n$($n$-1)/2).

Results obtained thus far cover only a small range of dimensionalities ($n$=5 and $n$=7), although this range is most common for criteria hierarchies used in AHP and related methods. Usage of this ordinal preference information in conjunction with (quasi-)regular minimum-diameter incomplete preference graphs seem to be a promising research direction, which can allow decision-makers to reduce the required number of comparisons required from experts while maintaining sufficient decision credibility levels. 

As all listed ranking-dependent PC patterns can be used in AHP and related methods, we consider both the current research and subsequent comparative experimental studies an important contribution to AHP theory and methodology, as well as to analysis of cognitive, algorithmic, and applied aspects of decision-making in general.

\newpage

\section*{Appendix. Flow-chart of the PC generation algorithm enabling us to compare max difference method with Best-worst and Top 2 methods.}
\label{sec:headings}

\begin{figure}[H]
    \centering
    \includegraphics[width=1\linewidth]{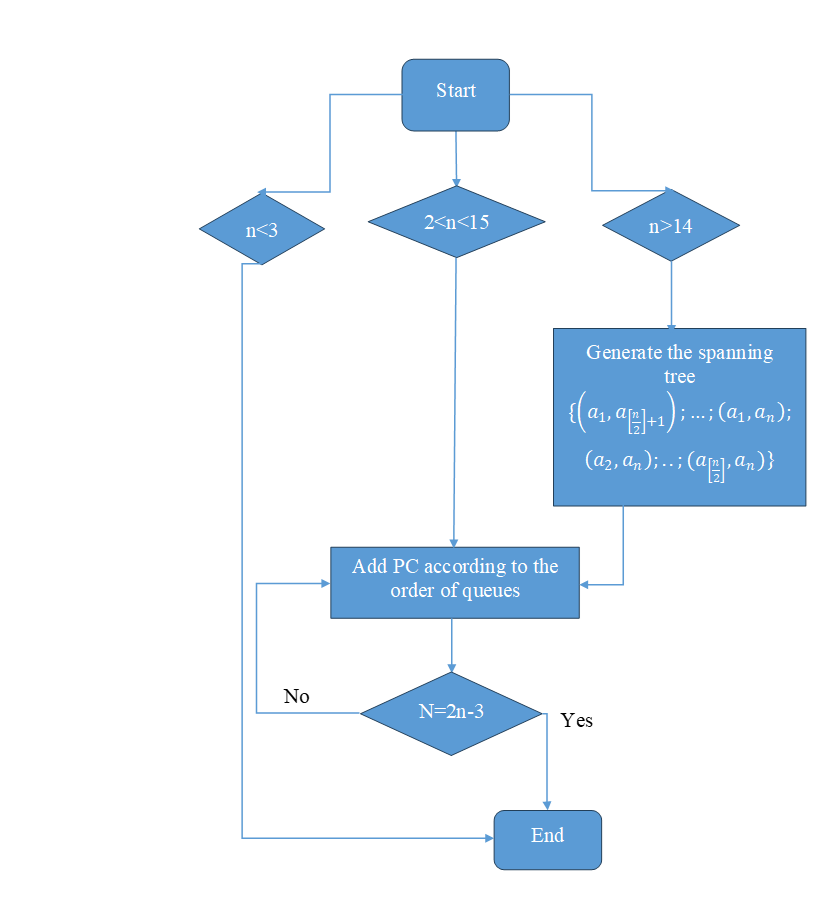}
%    \caption{Enter Caption}
    \label{fig:placeholder}
\end{figure}

\end{document}